\documentclass[conference]{IEEEtran}  
\IEEEoverridecommandlockouts

\newlength{\ConfHeaderTopLift}   
\newlength{\HeaderToTitleGap}    
\newlength{\TitleToAuthorGap}    

\makeatletter

\usepackage{siunitx}
\usepackage{tabularx} 

\usepackage{cite}
\usepackage{lipsum}
\usepackage{amsmath,amssymb,amsfonts}
\usepackage{algorithmic}
\usepackage{booktabs}
\usepackage{graphicx}
\usepackage{textcomp}
\usepackage{xcolor}
\usepackage{threeparttable}
\usepackage{graphicx} 
\usepackage{booktabs}
\usepackage{multirow}
\def\BibTeX{{\rm B\kern-.05em{\sc i\kern-.025em b}\kern-.08em
    T\kern-.1667em\lower.7ex\hbox{E}\kern-.125emX}}
\begin{document}


\title{MicCheck: Repurposing Off-the-Shelf Pin Microphones for Easy and Low-Cost Contact Sensing}

\author{Steven Oh$^{*}$, Tai Inui$^{*}$, Magdeline Kuan$^{*}$, Jia-Yeu Lin}


\makeatletter
\let\oldmaketitle\maketitle
\renewcommand{\maketitle}{%
  \oldmaketitle
  \thispagestyle{plain}%
  \renewcommand{\thefootnote}{}
    \footnotetext{%
      \vspace{0.5em}%
      \noindent\rule[0.5ex]{3cm}{0.4pt}\\[-0.5em]
      $^{*}$Equal contribution.
    }

  \renewcommand{\thefootnote}{\arabic{footnote}}
}
\makeatother

\maketitle

\begin{abstract}
Robotic manipulation tasks are contact-rich, yet most imitation learning (IL) approaches rely primarily on vision, which struggles to capture stiffness, roughness, slip, and other fine interaction cues. Tactile signals can address this gap, but existing sensors often require expensive, delicate, or integration-heavy hardware. In this work, we introduce MicCheck, a plug-and-play acoustic sensing approach that repurposes an off-the-shelf Bluetooth pin microphone as a low-cost contact sensor. The microphone clips into a 3D-printed gripper insert and streams audio via a standard USB receiver, requiring no custom electronics or drivers. Despite its simplicity, the microphone provides signals informative enough for both perception and control. In material classification, it achieves 92.9\% accuracy on a 10-class benchmark across four interaction types (tap, knock, slow press, drag). For manipulation, integrating pin microphone into an IL pipeline with open source hardware improves the success rate on picking and pouring task from 0.40 to 0.80 and enables reliable execution of contact-rich skills such as unplugging and sound-based sorting. Compared with high-resolution tactile sensors, pin microphones trade spatial detail for cost and ease of integration, offering a practical pathway for deploying acoustic contact sensing in low-cost robot setups. 
\end{abstract}

\section{Introduction}
Imitation learning (IL) has advanced robot manipulation substantially, yet many everyday skills remain contact–rich: task success often hinges on cues that are difficult to perceive with vision alone (e.g., stiffness, roughness, damping, incipient slip, and micro-impacts at the contact interface). Tactile and acoustic feedback can complement vision by sensing contact events directly and by improving robustness under occlusion or challenging illumination. However, practical adoption of tactile sensing faces a deployment gap: higher-performance solutions (e.g., vision-based tactile sensors, custom contact microphones, piezoelectric arrays) tend to be costly, fragile, or integration-intensive (amplifiers, drivers, bespoke software), limiting their use outside well-equipped laboratories. Notably, many manipulation tasks do not require ultra-fine spatial resolution; rather, they benefit from reliable, timely signals that separate no-contact from meaningful contact, discriminate broad material classes, and flag events such as slip or impact. In such regimes, a simpler, lower-cost, and easier-to-integrate sensor can be a reasonable trade-off.
\begin{figure}
    \centering
    \includegraphics[width=1.0\linewidth]{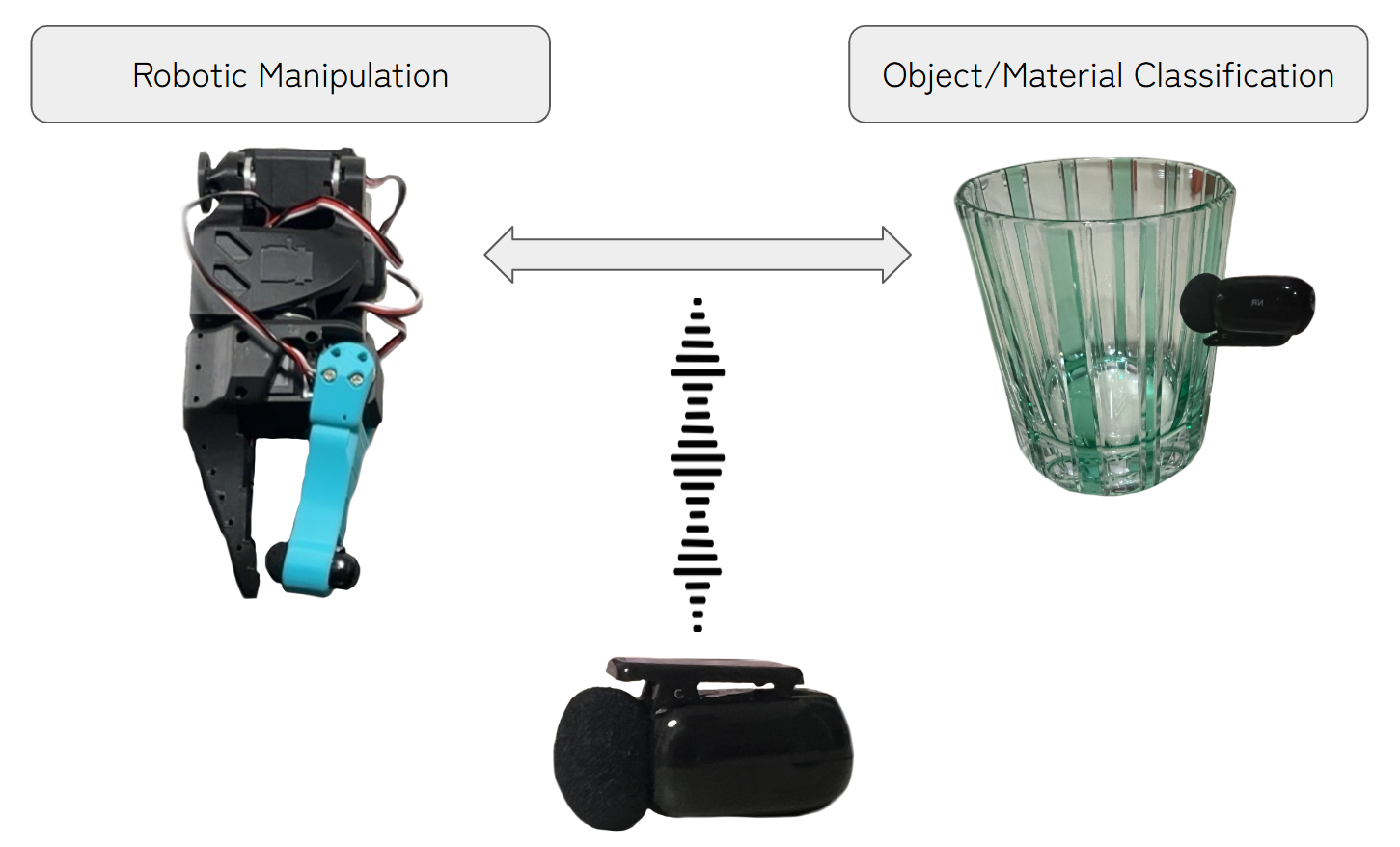}
    \caption{Overview of MicCheck. We repurpose low-cost pin microphones for contact sensing. We demonstrate this through two experiments: robotic manipulation and object classification.}

        \label{fig:teaser}
\end{figure}
We propose \emph{MicCheck}, a plug-and-play approach that repurposes an off-the-shelf Bluetooth pin microphone as a low-cost contact sensor. The microphone clips into a 3D-printed gripper insert and operates out of the box—no custom electronics or drivers—while the stock foam provides compliance and robust acoustic coupling. Despite its simplicity, MicCheck yields signals that are informative for both perception and control: in material classification, it attains 92.9\% window-level accuracy on a 10-class benchmark, and in manipulation, incorporating audio into an IL policy improves bottle-cap pouring success from 0.40 (vision only) to 0.80 (vision+audio) and supports two additional contact-rich tasks (texture sorting and high-friction unplugging). These results indicate that low-cost acoustic sensing can provide useful contact awareness, though it does not replace higher-resolution tactile sensors for fine spatial discrimination.

Our contributions are as follows (Fig. \ref{fig:teaser}): (i) A minimal, low-cost acoustic contact sensor using an unmodified consumer microphone and a simple mechanical mount; (ii) integration into classification and an IL pipeline showing consumer microphone complements imitation learning and is beneficial to contact-rich tasks.

In this paper, Sec.~\ref{sec:related} reviews tactile and acoustic sensing and positions our work. Sec.~\ref{sec:setup} details hardware, signal processing, and the learning setup. Sec.~\ref{sec:results} presents material classification and real-world manipulation results with ablations and discussion. Sec.~\ref{sec:conclusion} concludes, and Sec.~\ref{sec:limitations} outlines limitations and future directions.

\section{Related Work}
\label{sec:related}
\subsection{Passive Tactile Sensing}
\label{sec:related-passive}
Passive tactile sensing methods measure interaction signals without injecting external stimuli. Traditional approaches rely on resistive, capacitive, or force–torque sensors, while more recent systems leverage vision-based tactile designs. Notable examples include the TacTip family of optical biomimetic fingertips \cite{WardCherrier2018TacTip}, GelSight for high-resolution geometry and force estimation \cite{Yuan2017GelSight}, and derivatives such as GelSlim \cite{Donlon2018GelSlim}, DIGIT \cite{Lambeta2020DIGIT}, and OmniTact \cite{Padmanabha2020OmniTact}. Magnetic-based sensors such as ReSkin \cite{Bhirangi2021ReSkin} offer scalable, low-cost tactile skins. Passive acoustic sensing has also been explored, e.g., SonicSense \cite{liu2024sonicsense}, which embeds contact microphones into a multi-fingered hand to capture vibrations for object recognition and material classification. These methods demonstrate that passive signals can encode rich contact information, though often at the cost of custom fabrication or integration.

\subsection{Active Sensing}
\label{sec:related-active}
In contrast, active tactile sensing injects energy—through motion or actuation—into the environment and interprets the response. Active haptic perception surveys \cite{Seminara2019ActiveHapticsSurvey, Bajcsy2018RevisitingActivePerception} highlight how active strategies enable disambiguation of materials and shapes. Lepora et al. demonstrated exploratory tactile servoing using TacTip \cite{Lepora2017ExploratoryServoing}, while more recent work formalizes servoing with pose and shear features \cite{Lepora2024PoseShearServoing}. Martinez-Hernandez et al. \cite{MartinezHernandez2017ActiveSensorimotor} applied active sensorimotor control for autonomous tactile exploration, and Shahidzadeh et al. \cite{shahidzadeh2024actexploreactivetactileexploration} combined reinforcement learning with active tactile exploration for shape inference. Within acoustics, Lu and Culbertson \cite{Lu2023ActiveAcoustic} demonstrated active acoustic sensing for grasp state estimation, while VibeCheck \cite{Zhang2025VibeCheck} achieved robust peg-in-hole insertion using only an emitter–receiver acoustic pair. Active methods thus provide controllable, discriminative signals but increase hardware complexity.

\subsection{Acoustic Sensing in Robot Learning}
\label{sec:related-audio-rl}
Acoustic signals have recently been explored as a complementary modality for robot learning in contact-rich tasks. Some studies use audio–visual pretraining to improve generalization in low-data regimes \cite{mejia2024hearingtouch}, while others show that incorporating audio with vision enables better adaptation to texture changes, slip events, and hidden object states \cite{liu2024maniwav}. Multimodal systems that fuse vision, touch, and audio also improve performance on tasks such as dense packing and pouring by combining global, temporal, and local cues \cite{li2022seehearfeel}.

Beyond passive use, active acoustic sensing has been applied to infer material properties and grasp states through wave transmission \cite{Lu2023ActiveAcoustic}, and recent work shows that audio-only feedback can support robust imitation-learned peg-in-hole insertion \cite{Zhang2025VibeCheck}. Together, these results highlight how acoustic cues capture contact events that are hard to perceive visually or tactually. Our work follows this trend but emphasizes accessibility, using an off-the-shelf microphone as a plug-and-play solution for perception and imitation learning.

\begin{figure*}[t]
    \centering
    \includegraphics[width=1\linewidth]{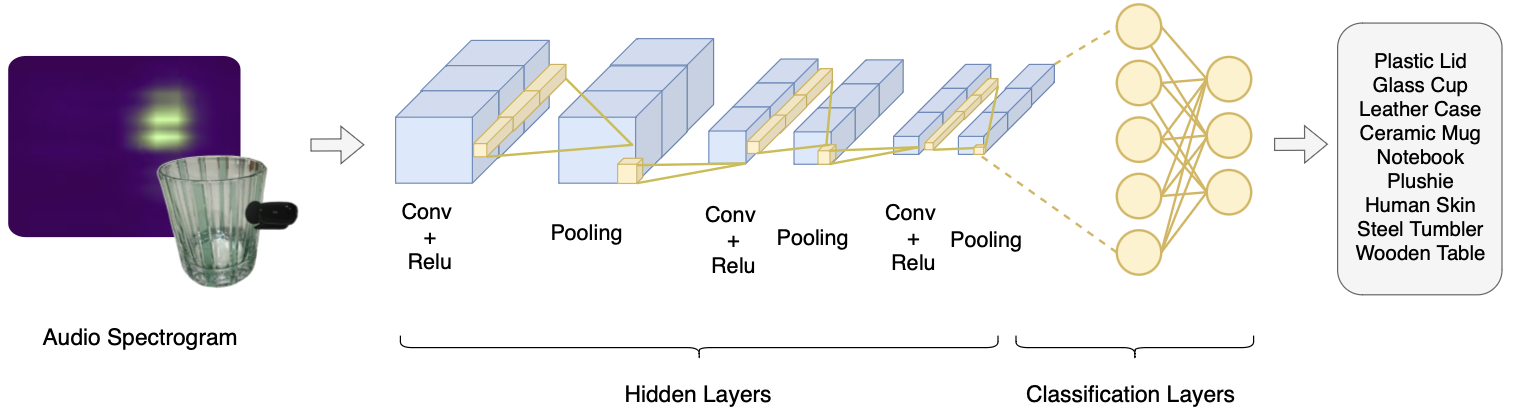}
    \caption{Architecture of the contact-based object classification model. 
Single-channel Mel spectrograms from 4 types of mic–object interactions 
(tap, knock, slow, drag) on 9 objects plus a ``blank'' no-contact class are 
fed into a compact 2D CNN comprising three Conv–BN–ReLU blocks, followed by 
global (adaptive) average pooling and a linear classifier. 
Models were trained with an 8:2 train/validation split (stratified by class) 
using cross-entropy loss and the Adam optimizer (learning rate $3\times10^{-4}$, 
batch size 32) for 2000 epochs, with the best checkpoint selected by highest 
validation accuracy. The blank class in the softmax serves as a rejection 
threshold for low-evidence windows.}
    \label{fig:class_sys}
\end{figure*}

\begin{figure}[t]
    \centering
    \includegraphics[width=1\linewidth]{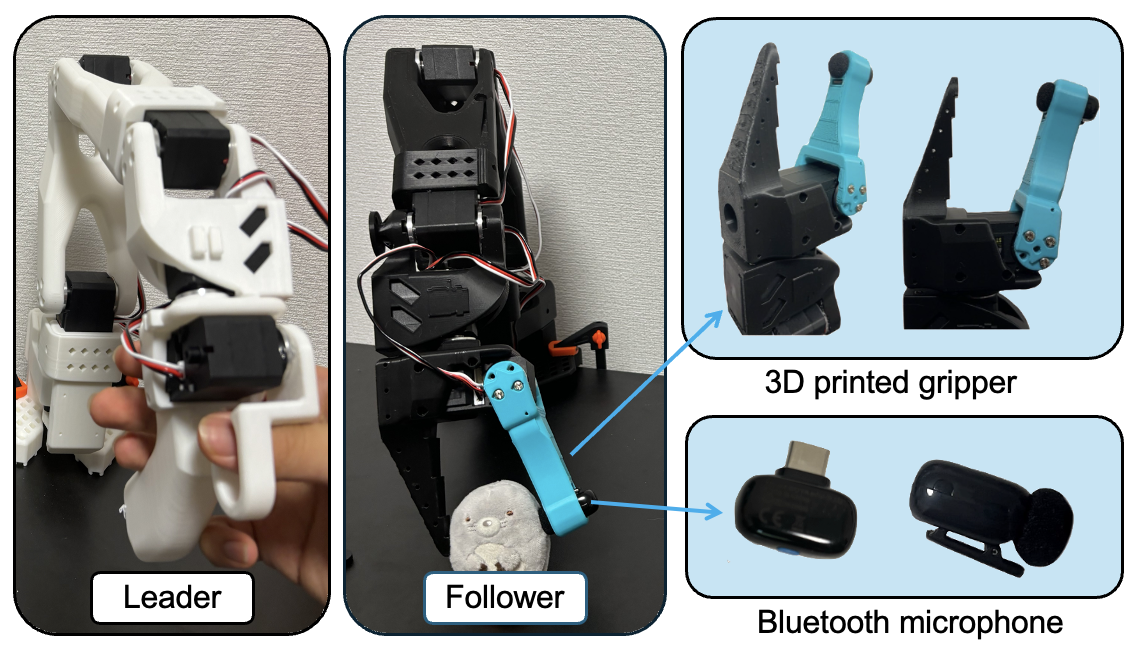}
    \caption{Teleoperation setup. We employ the Lerobot SO-101 teleoperation setup with a modified gripper. A commonly found commercial bluetooth microphone is embedded onto the gripper. The microphone is connected to a PC via a wireless USB retriever.}
    \label{fig:teleop}
\end{figure}

\begin{figure*}[t]
    \centering
    \includegraphics[width=1\linewidth]{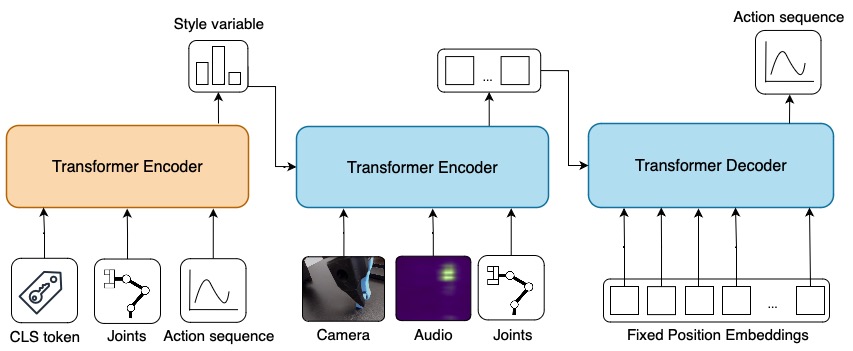}
\caption{Action Chunking with Transformers (ACT) architecture. Training uses a conditional variational autoencoder: a transformer episode/style encoder produces a latent $\mathbf{z}$ and a transformer encoder–decoder predicts a chunk of future actions conditioned on observations and $\mathbf{z}$. At inference, the transformer encoder is omitted to generate actions in fixed-size chunks.}

    \label{fig:act}
\end{figure*}

\section{Method}
\label{sec:setup}

\subsection{Experimental Setup}
We used an existing wireless pin microphone (BOYA mini, model \emph{mini-14}, 2.4\,GHz) as acoustic sensor. The clip-on transmitter is press-fit into the gripper mount so that its foam pad serves as the contact interface; the included USB-C receiver (dongle) plugs into the host PC and is enumerated as a standard audio input (Fig.~\ref{fig:teleop}). To suppress ambient and motor noise during motion, we enable the microphone’s built-in noise-cancellation. Audio is recorded at 48\,kHz/16-bit; we use a single transmitter in all experiments.

\subsection{Material Classification}
\label{sec:classification}
\subsubsection{Data Collection and Signal Processing}
\label{sec:classification-data}
We recorded contact sounds using the pin microphone mounted on the gripper’s foam face while interacting with 9 everyday objects spanning rigid solids (plastic lid, glass cup, ceramic mug, steel tumbler, wooden table) and compliant/textured surfaces (leather case, plush toy, notebook), plus a \emph{blank} (no-contact) condition. Speed and normal force were varied across trials.

We collected four interaction types with the selected objects:
\begin{itemize}
    \item \textbf{Tap}: a light, brief touch followed by immediate release.
    \item \textbf{Knock}: a firmer impact that tends to excite audible resonance.
    \item \textbf{Slow press}: a gradual press and hold with mainly normal contact.
    \item \textbf{Drag}: continuous sliding while maintaining contact.
\end{itemize}

Audio is converted to Mel-spectrogram features. Long recordings are segmented into fixed 1\,s windows without overlap; each window is mapped to a log-magnitude Mel spectrogram. This configuration captures resonant peaks, spectral envelopes, and decay characteristics informative for material and structure. Windows with no contact form the \emph{blank} class to avoid forced guesses in the network.

\subsubsection{Model}
\label{sec:classification-model}
We evaluate a compact 2D CNN on single-channel Mel spectrograms (Fig.~\ref{fig:class_sys}). The network comprises three Conv--BN--ReLU blocks, global adaptive average pooling, and a linear classifier. Windowed examples are split 8:2 (train/validation) with stratification by object class. Models are trained with cross-entropy using Adam (lr \(=3\!\times\!10^{-4}\), batch size \(=32\)) for 2000 epochs; the best checkpoint is selected by highest validation accuracy.

\begin{table}[t]
  \centering
  \footnotesize
  \caption{ACT Training Parameters}
  \label{tab:act_params}
  \begin{minipage}{1.0\linewidth}
    \centering
    \begin{tabular}{lc}
      \toprule
      \textbf{Parameter} & \textbf{Default} \\
      \midrule
      Chunk size       & 100 \\
      Backbone         & ResNet-18 \\
      Pretrained       & ImageNet-1K (ResNet18) \\
      Pre-norm         & False \\
      Model dim        & 512 \\
      Heads            & 8 \\
      FFN dim          & 3200 \\
      Activation       & ReLU \\
      Enc. layers      & 4 \\
      Dec. layers      & 1 \\
      VAE              & True \\
      Latent dim       & 32 \\
      VAE enc. layers  & 4 \\
      Dropout          & 0.1 \\
      KL weight        & 10.0 \\
      Learning rate    & \num{1e-5} \\
      Weight decay     & \num{1e-4} \\
      Backbone LR      & \num{1e-5} \\
      \bottomrule
    \end{tabular}
  \end{minipage}
\end{table}

\subsection{Robot Hardware and Teleoperation Setup}
\label{sec:robot-teleop}
We integrate a commercial pin microphone with the LeRobot SO101 platform \cite{cadene2024lerobot}. The robot gripper is redesigned to accept the microphone via its built-in clip; the mounting hole is dimensioned for a tight press-fit so the unit can be inserted/removed without additional fasteners. The microphone is oriented perpendicular to the gripper such that its foam pad becomes the primary contact surface on that side during interaction. This foam provides (i) compliant contact for stable grasping and (ii) effective acoustic coupling during object contact, enabling seamless audio capture without modifying the microphone form factor. We enable the device’s built-in noise-cancellation feature to reduce ambient and motor noise, particularly during motion.

For data collection, we use a teleoperation setup with a leader–follower configuration (Fig.~\ref{fig:teleop}). Joint states from the leader are streamed to the follower for, a strategy shown to be effective for collecting high-quality demonstrations for robot learning \cite{wu2023gello}. For training, each dataset consisted of 20 demonstrations per task, collected via teleoperation with synchronized RGB, proprioception, and audio. Compared to vision-only pipelines, the extra audio stream introduced negligible overhead. At inference, the multimodal pipeline ran at 50 Hz, bottlenecked by the camera framerate.

\subsection{Imitation Learning}
\label{sec:il}
To capture fine-grained contact dynamics, the audio stream is segmented into 0.2\,s frames at 30\,Hz (a new frame every 0.04\,s; 80\% overlap). Each frame is converted into a Mel spectrogram with $n_{\mathrm{mels}}{=}32$. Frequencies above $0.3\times$ the Nyquist rate are amplified by a factor of $2.0$ to emphasize sharp impact sounds. Compared to material classification, we adopt shorter and more overlapping windows here to ensure transient acoustic events remain temporally aligned with proprioceptive and visual signals.

We train an Action Chunking with Transformers (ACT) policy~\cite{zhao2023learningfinegrainedbimanualmanipulation} that predicts fixed-length action sequences conditioned on recent multimodal observations (Fig.~\ref{fig:act}). ACT is an imitation learning approach that mitigates compounding error in sequential control by predicting a chunk of future actions at each step, rather than a single next action. This chunked prediction reduces the effective task horizon and improves the smoothness and robustness of the learned behavior.  

In our implementation, observations include (i) RGB images from a stationary camera, (ii) the most recent audio spectrogram frame, and (iii) robot proprioceptive states. These modalities are fused into a unified embedding and processed by a transformer encoder–decoder that outputs target joint positions for the next $H$ timesteps. The model is trained for 100k steps using the hyperparameters listed in Tab.~\ref{tab:act_params}.

\begin{figure}
    \centering
    \includegraphics[width=1.0\linewidth]{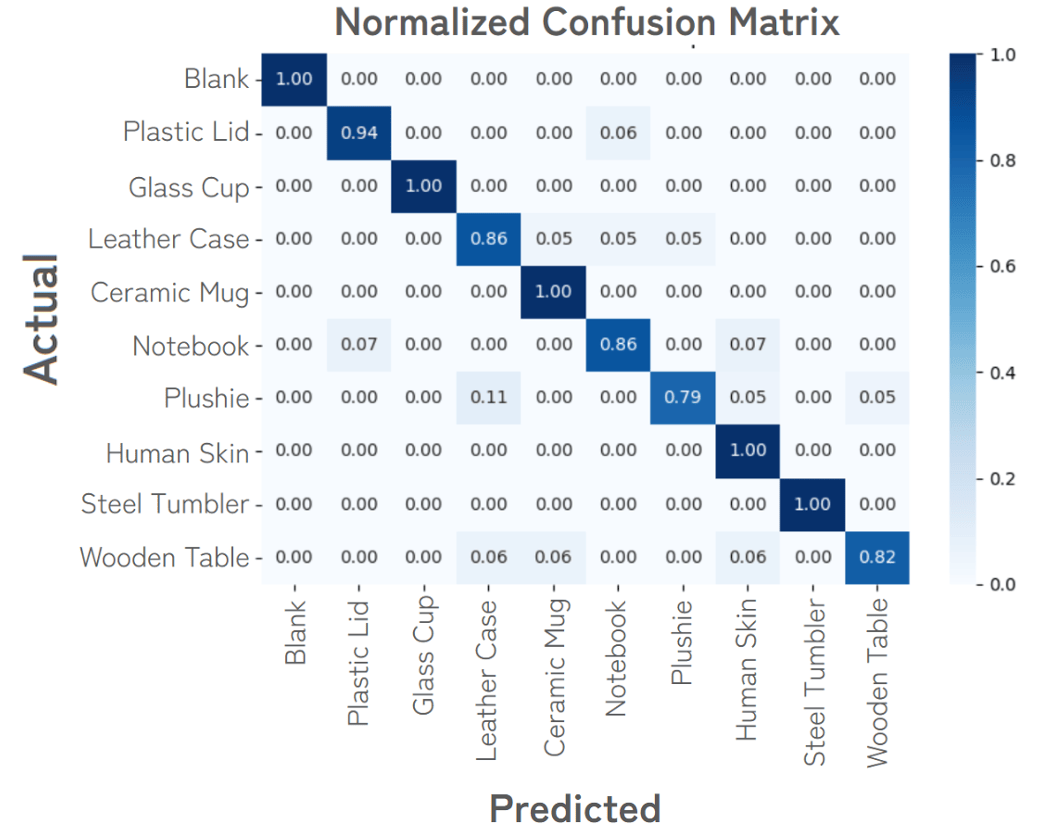}
\caption{Normalized confusion matrix for the 10-class (9 objects + ``blank'') 
material classification task. The model shows strong diagonal dominance, with perfect 
accuracy for the blank class, glass cup, human skin, and steel tumbler. Most 
confusions occur between acoustically similar soft materials (e.g., plushie vs.\ leather 
case, notebook vs.\ leather case), reflecting challenges in distinguishing objects 
with overlapping frequency responses.}

        \label{fig:confmat}
\end{figure}

\section{Results}
\label{sec:results}
\subsection{Material Classification}
\label{sec:results-classification}
We obtain 92.9\% window-level classification accuracy on the 9 objects plus a \emph{blank} class. The confusion matrix in Fig.~\ref{fig:confmat} shows strong diagonal dominance, with perfect separation for \emph{blank}, glass cup, ceramic mug, human skin, and steel tumbler, and a high score for plastic lid (0.94). Inaccuracies concentrate in the softer/textured group: plushie is sometimes predicted as leather (0.11) and, to a lesser extent, notebook; leather and notebook also show small mutual bleed while each remains at 0.86 on the diagonal. Wooden table is somewhat less distinct (0.82 on-diagonal), with small leaks into nearby rigid classes (e.g., 0.06 to ceramic). Cross-family errors are rare (rigid items are seldom mistaken for soft ones), and the clean "blank" row/column indicates reliable no-contact rejection. Overall, errors arise mainly among classes with acoustically similar signatures (soft materials and wood) rather than across clearly different materials. 

These results indicate that a single, low-cost pin microphone can achieve material discrimination with reasonably high accuracy across a range of object types and interactions. Most misclassifications occur between acoustically similar soft or textured objects, suggesting that the sensor captures the dominant frequency features but may struggle with fine-grained distinctions. The reliable rejection of the \emph{blank} class shows that the system can consistently separate contact from no-contact events, an essential property for integration into robot control pipelines. Overall, the analysis suggests that low-cost pin microphone provides a simple and effective way to add contact awareness across materials and contact types, without needing the resolution of more advanced tactile sensors.

\begin{figure*}[t]
    \centering
    \includegraphics[width=\linewidth]{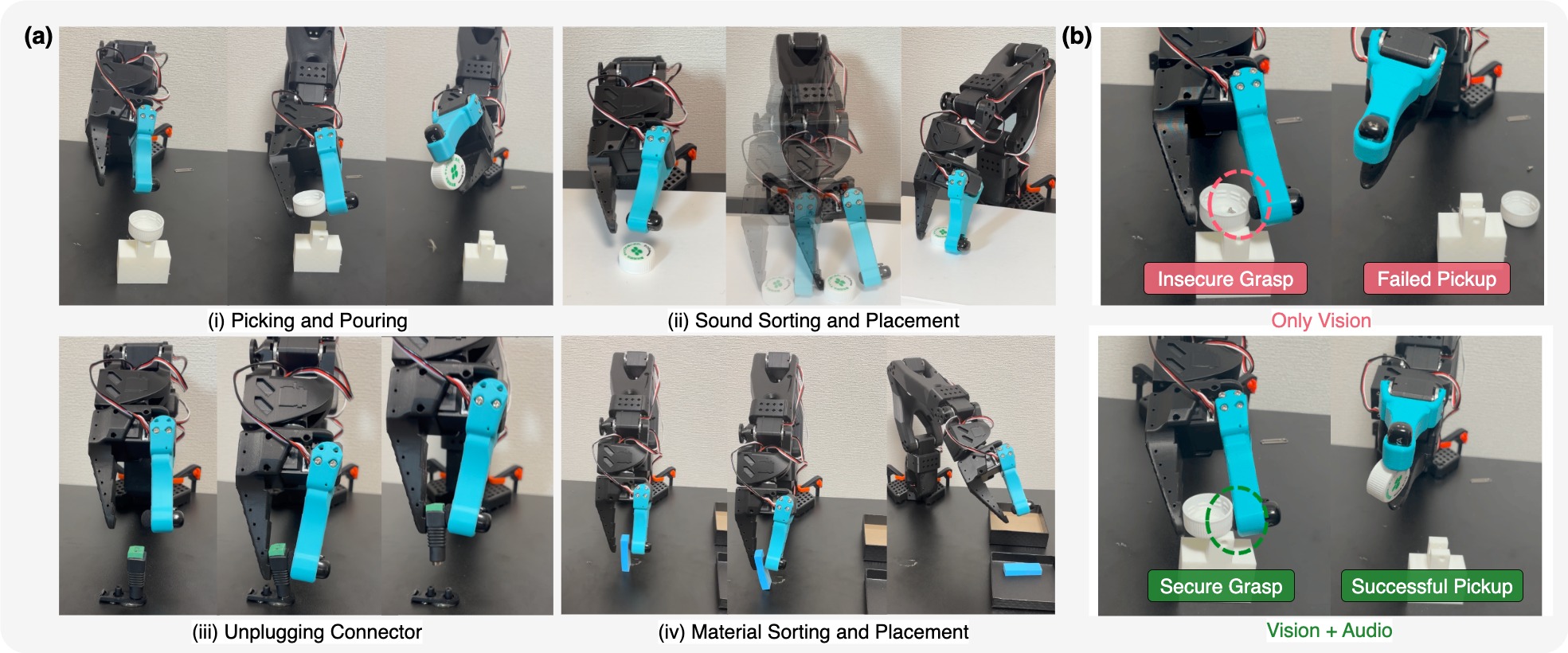}
    \caption{(a) Demonstration of four manipulation tasks: (i) Picking and pouring — the robot picks up a plastic bottle cap and pours a metallic screw inside; (ii) Sound-based sorting and placement — the robot shakes the cap and sorts it left or right depending on whether a metallic screw is detected acoustically; (iii) Unplugging connector — removing a tightly seated connector that requires high frictional force; and (iv) Material sorting and placement — distinguishing between bare sticky-note pads and ones covered with a plastic film, and sorting them into separate containers.
(b) Ablation study on sensing modality tested on task (a-i). With vision-only input, the policy often results in insecure grasps and dropped caps due to the cap’s deformability. Incorporating microphone (vision + audio) feedback enables the policy to detect contact states acoustically, achieving more secure and consistent grasps.}
    \label{fig:real-world-task}
\end{figure*}

\begin{table}[t]
\centering
\caption{Real world task performance}
\label{tab:combined_perf}
\setlength{\tabcolsep}{5pt}  
\renewcommand{\arraystretch}{1.05}  
\begin{tabular}{@{}lll c@{}}
\toprule
\textbf{Task} & \textbf{Condition} & \textbf{Input} & \textbf{Success} \\
              &                    & \textbf{modality} & \textbf{rate} \\
\midrule
(i) Picking and Pouring & \textemdash & Vision only    & 0.40 \\
                        & \textemdash & Vision + Audio & 0.80 \\
\midrule
(ii) Sound Sorting & Has Sound & Vision + Audio & 0.70 \\
                   & No Sound  & Vision + Audio & 0.60 \\
\midrule
(iii) Unplugging Connector & \textemdash & Vision + Audio & 1.00 \\
\midrule
(iv) Material Sorting & Plastic End & Vision + Audio & 0.70 \\
                      & Normal End  & Vision + Audio & 0.40 \\
\bottomrule
\end{tabular}
\end{table}

\subsection{Imitation Learning in Real-World Tasks}
\label{sec:results-il}
We first quantify the contribution of audio by ablating sensory inputs on a picking and pouring task (see Fig. ~\ref{fig:real-world-task} a(i)). We use the same dataset but removing the audio tokens from the input at training. As summarized in Table~\ref{tab:combined_perf}, success improves from 0.40 with vision only to 0.80 with vision+audio over 10 roll-outs per setting. Fig.~\ref{fig:real-world-task}b illustrates this qualitative difference: the vision-only policy often slips and fails to complete the pour, whereas adding audio yields a stable grasp and consistent rotation aligned with demonstrations.

To showcase versatility beyond this ablation, we evaluate three additional contact-rich tasks (Fig.~\ref{fig:real-world-task}). Table~\ref{tab:combined_perf} reports their success: unplugging achieves 1.00; sorting by sound reaches 0.70 (has object) and 0.60 (no object); and sorting by texture achieves 0.70 (plastic end) and 0.40 (normal end). Each policy was trained on 20 demonstrations per task (10 per condition for binary tasks) and is evaluated with 10 roll-outs per condition.

Failures primarily stem from insufficient contact leading to aborted attempts, or from incorrect sorting decisions caused by ambient or motor-induced noise. Overall, audio cues proved most reliable when contact interactions generated salient and repeatable sound signatures (e.g., during unplugging). In addition to providing tactile-like feedback through sound, the soft foam surrounding the pin microphone also functioned as a compliant contact interface, improving grip stability and sound coupling during manipulation.

These results show the practicality and value of incorporating a simple commercial microphone for robot learning. Beyond providing complementary acoustic information that vision alone cannot capture, the microphone itself serves as a low-cost and easily integrable sensor that inherently adds compliance through its soft foam housing. Such simplicity lowers the barrier for deploying multimodal perception in everyday robotic systems, making it feasible to scale imitation learning beyond specialized research settings.

\section{Limitations and Future Work}
\label{sec:limitations}
While our results show that an off-the-shelf pin microphone can provide useful contact information, several limitations remain. The current system relies on a single consumer-grade microphone with wireless audio, which introduces compression artifacts, latency, and occasional noise contamination. Our experiments also focused on a modest set of objects and interactions, meaning results may not fully generalize to broader manipulation tasks.

Future work should therefore pursue a more rigorous evaluation. This includes comprehensive ablation studies to isolate the role of audio relative to vision and proprioception, including different encoding methods, and expanded benchmarks with larger, more diverse objects and contact types. Furthermore, noise reduction by using pretrained dataset or learning-based approach could improve noise to signal ratio, which needs to be further benchmarked against existing tactile sensors. Additionally, multi-microphone configurations and improved placement strategies using other low-cost sensors could further extend sensing range beyond single-point contact.
\section{Conclusion}
\label{sec:conclusion}

This work explored a minimal and accessible approach to contact sensing for robot learning by repurposing an off-the-shelf Bluetooth pin microphone as an acoustic tactile sensor. We demonstrated that such a simple, low-cost device—integrated via a 3D-printed mount and used without any custom electronics—can produce informative signals for both perception and control. Despite its hardware simplicity, the system achieved high material classification accuracy and improved the robustness of imitation-learned manipulation policies in contact-rich settings. While the performance does not match that of high-resolution tactile sensors, our findings suggest that audio can serve as a lightweight complementary modality to vision and proprioception, offering meaningful cues about contact state, material type, and interaction dynamics. This trade-off between fidelity and accessibility highlights a promising direction for scaling multimodal robot learning beyond specialized laboratory setups.

Overall, we show that commodity microphones, when thoughtfully integrated, can extend the sensory capabilities of everyday robots. We view this not as a replacement for tactile sensors, but as a practical step toward democratizing multimodal sensing—enabling broader experimentation, reproducibility, and adoption of contact-aware learning systems in the robotics community.

\section*{Acknowledgment}
This work was supported by the JSPS Grant-in-Aid for Early-Career Scientists [23K12755].
\clearpage

\bibliographystyle{IEEEtran}
\bibliography{references}

\end{document}